# Semantics-Driven Cloud-Edge Collaborative Inference

## A Case Study of License Plate Detection

Yuche Gao

Department of Statistical Science

University College London

London, United Kingdom

yuchegao@gmail.com

Beibei Zhang

Intelligent Software Research Center

Zhejiang Lab

Hangzhou, China

beibei@zhejianglab.com

*Abstract*—With the proliferation of video data in smart city applications like intelligent transportation, efficient video analytics has become crucial but also challenging. This paper proposes a semantics-driven cloud-edge collaborative approach for accelerating video inference, using license plate recognition as a case study. The method separates semantics extraction and recognition, allowing edge servers to only extract visual semantics (license plate patches) from video frames and offload computation-intensive recognition to the cloud or neighboring edges based on load. This segmented processing coupled with a load-aware work distribution strategy aims to reduce end-to-end latency and improve throughput. Experiments demonstrate significant improvements in end-to-end inference speed (up to 5x faster), throughput (up to 9 FPS), and reduced traffic volumes (50% less) compared to cloud-only or edge-only processing, validating the efficiency of the proposed approach. The cloud-edge collaborative framework with semantics-driven work partitioning provides a promising solution for scaling video analytics in smart cities.

## I. INTRODUCTION

Intelligent transportation is a crucial domain in the urban brain [1]. It utilizes modern information technologies to conduct intelligent management and planning of urban traffic, improving transportation efficiency, reducing congestion, enhancing safety, and raising service levels. Video systems are a key component of intelligent transportation, used for monitoring road traffic, vehicle movements, traffic violations, etc., providing data support for smart transportation. For example, in an urban brain smart transportation monitoring system, cameras can monitor urban intersections and highways in real time, analyzing video streams for optimal monitoring.

However, due to high road density and vehicle volume in cities, fully monitoring urban transportation requires numerous cameras, generating massive video data streams around the clock. Efficiently processing such vast amounts of video data poses huge challenges for back-end information systems. Directly sending video streams to the cloud for processing incurs massive communication and computational overheads. Therefore, cloud-edge collaboration (Fig. 1) has become a common approach for handling massive video data [2]. Edge servers collect video streams from multiple nearby cameras for real-time processing, then send processed results (e.g. vehicle volume, tracking info, pedestrian volume) to the cloud, significantly reducing communication costs and cloud server workloads.

In an urban brain context, edge servers are numerous. Controlling their configurations is key to reducing overall system costs [3]. Therefore, edge servers may become overloaded and unable to process data in real time when bursts of high traffic volumes occur, causing business interruptions. To prevent this, video processing workloads can be shifted to neighboring edge servers or the cloud. However, as modern video streams are typically 1080P or 2160P (4K) high definition, a 1080P video stream has a bitrate of 3500kbps. If an edge server connects to 8 cameras, the traffic totals 28Mbps. Whether redirected to the cloud or neighboring edges, such workloads impose huge burdens on communication networks and target compute nodes.

The key contributions of this paper include: 1) With license plate recognition, the most common scenario in smart transportation of city brains, as an example, comparing common license plate detection and recognition methods and proposing a semantics-driven edge-cloud collaborative license plate detection method. 2) Proposing an edge-cloud collaborative queue processing mechanism with congestion resistance based on the producer-consumer model. 3) Demonstrating the performance improvements of the proposed method in terms of end-to-end latency, throughput, edge-cloud traffic, device utilization and other evaluation metrics.

## II. Related Work

The existing methods are based on fast object detection in general video streams, which improve efficiency while ensuring accuracy. They mainly include detection and tracking combined video object detection, video stream object detection based on motion information feature migration or fusion, etc.

Detection and tracking combined video object detection is a common video object detection method. Its basic idea is to first detect targets in each frame of the video as static image object detection, and then use multi-target tracking algorithms to track the target boxes and use the tracking results to correct the previous detection results to improve stability and continuity. The advantage of this method is that it can utilize existing single-frame object detectors and multi-target trackers without the need to design complex network structures or training processes. The disadvantage of this method is that it relies on the performance of single-frame object detectors and multi-target trackers. If one of them fails, it will affect the effect of the entire video object detection. A representative work is T-CNN [4], which proposes a video object detection framework based on tracking and regression. It first uses Faster R-CNN [5] to detect targets in each frame of the video, then uses MDNet for multi-target tracking of the detection results, and finally uses a regression network to optimize and reorder the tracking results.

Video stream object detection based on motion information feature migration or fusion is a method that uses optical flow and other motion information to estimate feature changes between adjacent frames, and then transfers or fuses features from key frames to other frames to reduce duplicate calculations and improve consistency [6]. The advantage of this method is that it can use motion information to enhance the spatial and temporal information in the video to improve the accuracy and robustness of object detection. The disadvantage is that it requires additional computation of optical flow and other motion information, which increases the amount of computation and time overhead. A representative work is Deep Feature Flow [7], which proposes an optical flow based feature propagation method that maps features from key frames to other frames through optical flow, then fuses the propagated features and current frame features with a fusion module, and finally performs object detection with a detection module.

Although existing high performance encoding technologies represented by H.264 [8] can compress over 100:1, existing technologies do not effectively detect semantic information combined with business scenarios. In cloud-edge collaborative scenarios, it is necessary to transmit the complete video stream to adjacent nodes and use video object detection methods for repetitive processing. For edge scenarios of city brains, since high-definition cameras are used, the traffic is large, and edge servers are often cost-sensitive and have limited processing capabilities. In the face of sudden large traffic and sudden events, etc., processing delays may occur, leading to business flow interruptions and other events.

## III. System Model and Algorithm

As Fig. 1 shows, an edge license plate recognition system typically comprises: 1) Cameras, 2) Edge servers, and 3) Cloud servers. Cameras capture and push video streams to edge servers.

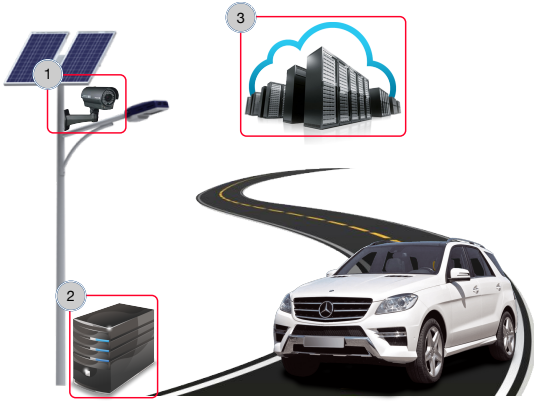

Figure 1: Edge license plate recognition system

Upon receiving streams, edge servers extract video frames for processing, then return results to the cloud for archiving/analysis.

License plate detection is a key smart transportation technology, used to automatically identify and extract plate information from vehicle images, accurately locating and recognizing plates in complex scenarios to provide important data support for transportation management, vehicle tracking, security monitoring, etc.

Typically, a license plate detection algorithm involves the following main steps:

1. Image preprocessing: Perform preprocessing operations like enhancement, de-noising, edge detection, etc. In the preprocess of a frame of video, the filtering-based methods such as median filtering and adaptive Wiener filtering can be used on input vehicle images to improve subsequent plate localization accuracy and robustness.
2. Plate localization: Leveraging various image processing techniques and feature analysis methods, accurately locate the plate region in a preprocessed vehicle image. This may involve color, shape, edge and other types of analysis.
3. Plate segmentation: Further process the localized plate region to segment out each character/digit, typically involving character spacing analysis, projection, template matching and other techniques to obtain clear character images.
4. Character recognition: Recognize and categorize the segmented character images into their corresponding characters/digits. This can be done using conventional machine learning like pattern recognition and feature extraction, or deep learning techniques like Convolutional Neural Networks (CNNs).
5. Post-processing: Correct and validate recognition results using rules and algorithms to eliminate errors and improve accuracy and reliability.

**Algorithm 1:** Patch Queue

1: $Q \leftarrow Queue(), \{p_1, p_2, \ldots p_{k-1}\}$; /* a queue of extracted license plate image patches waiting for character recognition */
2: **while** TRUE:
3:     $p_k \leftarrow$ The segmented license plate image;
4:     **if** $p_k$ is not Null: Q.append($p_k$);

**Algorithm 2:** Proposed Cloud-Edge Collaborative Algorithm

1: $Q \leftarrow Queue(), \{p_1, p_2, \ldots p_{k-1}\}$; /* a queue of extracted license plate image patches waiting for character recognition */
2: $N_{max} \leftarrow$ the maximum capacity of the local edge server;
3: send signals to check its neighbors;
4: **while** TRUE:
5:     **if** $|Q| > N_{max}$:
6:         **if** local edge is available:
7:             local edge recognizes characters and reports to cloud;
8:         **else if** any neighboring edge is available:
9:             send the head to the neighboring edge;
10:             neighboring edge recognizes characters and reports to cloud;
11:         **else**:
12:             send the head to cloud;
13:             cloud conducts recognition;
14:     **else if** $|Q| \leq N_{max}$ and $|Q| > 0$:
15:         local edge recognizes characters and reports to cloud;
16:     Q.remove(); /* remove the head which is processed */

We refer to the first three steps as semantics extraction, and the last two as semantics recognition. For a video stream's continuous frames, the edge server places each frame into a task queue (FIFO queue) forming a frame queue $N = \{n_1, n_2, \ldots, n_k\}$ and performs the five steps on each queued frame. However, the edge server has limited processing capacity. When bursts occur, queued frames ($|N| = k$) may exceed the queue's maximum capacity ($N_{max}$), i.e. $|N| > N_{max}$. To address this, we propose a segmented processing approach that separates semantics extraction and recognition, using the edge server to process local streams for semantics extraction. The extracted license plate image patches are packaged into objects $p_i$ to form a patch queue $Q = \{p_1, p_2, ... p_k\}$ (Algorithm 1). Then, for character recognition, when the length $|Q| = k$ is smaller than or equal to the maximum capacity, i.e. $|N| \leq N_{max}$, the local edge server works as the consumer, processing from the head. Or when the length ($|Q| = k$) exceeds $N_{max}$, the second consumer begins working for collaborative processing. It distributes patches at the head to neighboring edges, if available, or to the cloud, if neighboring edges are also overloaded. Finally, extracted plate fields are reported to the cloud (Algorithm 2).

## IV. EXPERIMENTAL RESULTS

Our edge nodes were low-power Intel i7-10510U 1.8GHz CPUs with 16GB RAM. The cloud server had an NVIDIA RTX 2080Ti GPU. Experiments used the standardized CCPD (Chinese City Parking Dataset) and video streams from a campus parking lot. Three combinations of semantics extraction and recognition algorithms were tested (Table 1). *A*. End-to-end latency, *B*. throughput, *C*. cloud-edge traffic, and *D*. device utilization were measured and compared between direct edge processing and our approach. The following results indicate improvements in all metrics:

A. latency decreased for all cases using our approach, by up to 1/5 of the original (Fig. 2).

B. Average throughput increased 5x with our approach, reaching ~9 frames per second (FPS) with Hyperlpr, ~8 FPS with YoLo, and ~9 FPS with MTCNN (Fig. 3).

C. With traffic bursts requiring cloud-edge collaboration, semantics-extracted patches were sent to other edges/cloud for processing. Compared to directly transferring unprocessed frames, total data transfer (measured in KB) on data links decreased in all cases. With Hyperlpr it averaged 845.22KB (51.81% of original). With YoLo it was 850.6KB (52.14% original). With MTCNN it was 841.15KB (51.57% original) (Fig. 4).

Table 1: License plate recognition algorithms

| Algorithm Group | Semantics Extraction | Semantics Recognition |
|---|---|---|
| Hyperlpr [9] | Mobilene-ssd | CTC |
| YoLo [10] | YoLov5 | CRNN |
| MTCNN [11] | MTCNN | LPRNET |

Figure 2: End-to-end inference latency

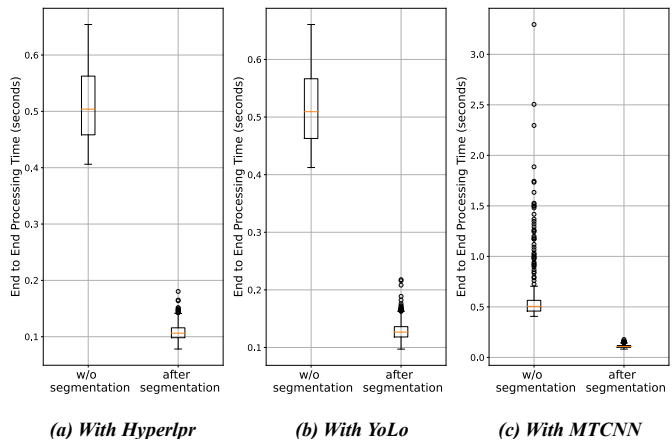

*(a) With Hyperlpr*   *(b) With YoLo*   *(c) With MTCNN*

Figure 3: Throughput

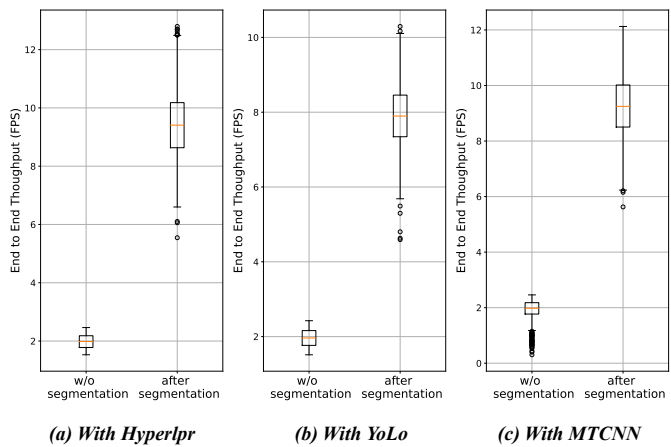

*(a) With Hyperlpr*   *(b) With YoLo*   *(c) With MTCNN*

Figure 4: Data Transfer

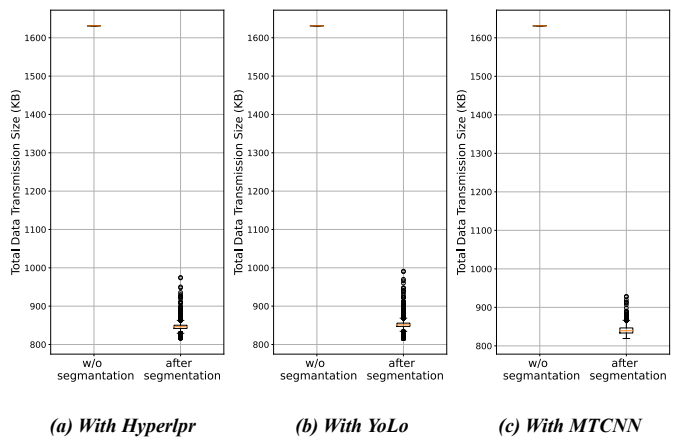

*(a) With Hyperlpr*   *(b) With YoLo*   *(c) With MTCNN*

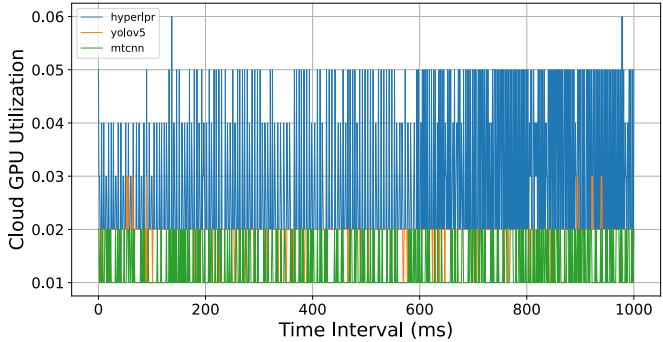

Figure 6: Cloud GPU Usage

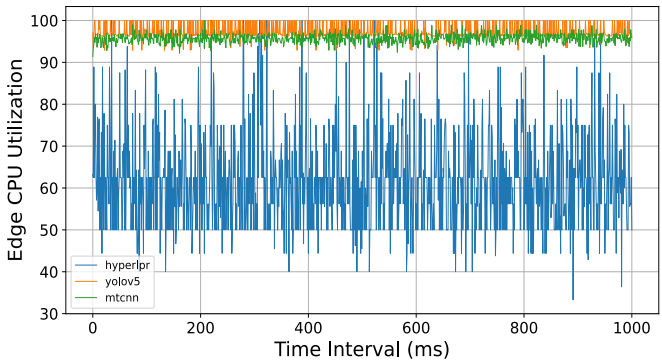

Figure 5: Edge CPU Usage

D. For device Utilization, we evaluated edge CPU and cloud GPU usage after applying our algorithm, measuring cloud GPU usage when sending patches there for semantics recognition (Fig.5 and Fig.6).

License Plate Recognition from Campus Video Streams: The above method was applied to campus video streams, accelerating license plate recognition as shown in Fig. 7.

Figure 7: Example of application

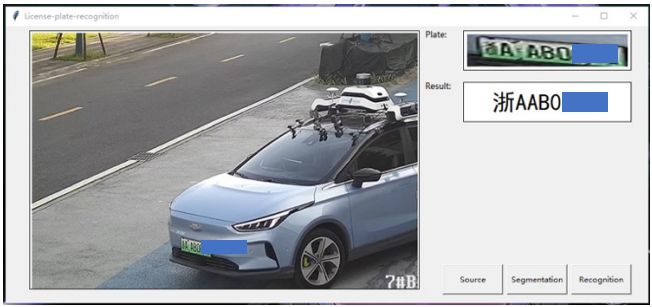

## V. CONCLUSION

In conclusion, we propose a novel semantics-driven cloud-edge collaborative approach to accelerate video analytics for smart city applications. Using license plate recognition as a case study, we develop a segmented processing methodology that separates semantics extraction on the edge and computation-intensive recognition on the cloud/edge. This allows us to efficiently extract visual semantics at the edge before intelligently distributing recognition workloads based on server loads. Comprehensive experiments demonstrate that our approach can significantly reduce end-to-end latency, improve throughput, decrease traffic volumes, and better utilize devices. The collaborative framework provides a scalable and efficient solution to meet the growing demands of real-time video analytics for smart transportation and other domains. While focusing on license plate recognition here, the semantics-driven methodology can be extended to other smart city scenarios. Further research can be done to explore more advanced semantics extraction and optimal work distribution strategies.